\documentclass{article}
\usepackage[utf8]{inputenc}

\title{Deep Reinforcement Learning and\\ its Neuroscientific Implications}

\author{{\bf Matthew Botvinick$^{1,2}$, Jane X. Wang$^{1}$} \\
        {\bf Will Dabney$^{1}$, Kevin J. Miller$^{1,2}$}, 
        {\bf Zeb Kurth-Nelson$^{1,2}$} \\
\vspace{0.15 cm}\\
$^1$DeepMind, UK, 
$^2$University College London, UK
}

\usepackage{subfig}
\usepackage[a4paper, margin=1in]{geometry}
\usepackage[T1]{fontenc}

\usepackage{natbib}
\usepackage{graphicx}
\usepackage{amssymb, amsmath, amsthm}

\usepackage{amsthm,amsmath,amssymb}
\usepackage{mathrsfs}

\usepackage{enumitem}
\setlist[itemize]{noitemsep}
\setlist[enumerate]{noitemsep}
\usepackage{hyperref}
\usepackage[dvipsnames]{xcolor}

\usepackage{algorithm}
\usepackage{algorithmic}
\usepackage{bbm}

\usepackage{verbatim}
\usepackage{graphicx}

\usepackage{thmtools}
\usepackage{thm-restate}

\begin{document}

\maketitle

\begin{abstract} 

The emergence of powerful artificial intelligence is defining new research directions in neuroscience. To date, this research has focused largely on deep neural networks trained using supervised learning, in tasks such as image classification. However, there is another area of recent AI work which has so far received less attention from neuroscientists, but which may have profound neuroscientific implications: deep reinforcement learning. Deep RL offers a comprehensive framework for studying the interplay among learning, representation and decision-making, offering to the brain sciences a new set of research tools and a wide range of novel hypotheses. In the present review, we provide a high-level introduction to deep RL, discuss some of its initial applications to  neuroscience, and survey its wider implications for research on brain and behavior, concluding with a list of opportunities for next-stage research.

\end{abstract}

\medskip

\section*{Introduction}

The last few years have seen a burst of interest in deep learning as a basis for modeling brain function  \citep{cichy2019deep, gucclu2017modeling, hasson2020direct, marblestone2016toward, richards2019deep}. Deep learning has been studied for modeling numerous systems, including vision \citep{yamins2014performance, yamins2016using}, audition \citep{kell2018task}, motor control \citep{merel2019hierarchical,weinstein2017structure}, navigation \citep{banino2018vector,whittington2019tolman} and cognitive control \citep{mante2013context, botvinick2014computational}.
This resurgence of interest in deep learning has been catalyzed by recent dramatic advances in machine learning and artificial intelligence (AI). Of particular relevance is progress in training deep learning systems using supervised learning – that is, explicitly providing the ‘correct answers’ during task training – on tasks such as image classification \citep{krizhevsky2012imagenet, deng2009imagenet}. 

For all their freshness, the recent neuroscience applications of supervised deep learning can actually be seen as returning to a thread of research stretching back to the 1980s, when the first neuroscience applications of supervised deep learning began \citep{zipser1988back,zipser1991recurrent}. Of course this return is highly justified, given new opportunities that are presented by the availability of more powerful computers, allowing scaling of supervised deep learning systems to much more interesting datasets and tasks. However, at the same time, there are other developments in recent AI research that are more fundamentally novel, and which have received less notice from neuroscientists. Our purpose in this article is to call attention to one such area which has vital implications for neuroscience, namely \emph{deep reinforcement learning} (deep RL). 
 
As we will detail, deep RL brings deep learning together with a second computational framework that has already had a substantial impact on neuroscience research: reinforcement learning. Although integrating RL with deep learning has been a long-standing aspiration in AI, it is only in very recent years that this integration has borne fruit. This engineering breakthrough has, in turn, brought to the fore a wide range of computational issues which do not arise within either deep learning or RL alone. Many of these relate in interesting ways to key aspects of brain function, presenting a range of inviting opportunities for neuroscientific research -- opportunities that have so far been little explored. 
 
In what follows, we start with a brief conceptual and historical introduction to deep RL, and discuss why it is potentially important for neuroscience. We then highlight a few studies that have begun to explore the relationship between deep RL and brain function. Finally, we lay out a set of broad topics where deep RL may provide new leverage for neuroscience, closing with a set of caveats and open challenges. 

\section*{An introduction to Deep RL}
 
\subsection*{Reinforcement learning}

Reinforcement learning \citep{sutton2018reinforcement} considers the problem of a learner or agent embedded in an environment, where the agent must progressively improve the actions it selects in response to each environmental situation or state (Figure ~\ref{fig:deep-rl}a). Critically, in contrast to supervised learning, the agent does not receive explicit feedback directly indicating correct actions. Instead, each action elicits a signal of associated reward or lack of reward, and the reinforcement learning problem is to progressively update behavior so as to maximize the reward accumulated over time. Because the agent is not told directly what to do, it must \textit{explore} alternative actions, accumulating information about the outcomes they produce, thereby gradually honing in on a reward-maximizing behavioral policy. 

Note that RL is defined in terms of the learning problem, rather than by the architecture of the learning system or the learning algorithm itself. Indeed, a wide variety of architectures and algorithms have been developed, spanning a range of assumptions concerning what quantities are represented, how these are updated based on experience, and how decisions are made.  

Fundamental to any solution of an RL problem is the question of how the state of the environment should be represented. Early work on RL involved simple environments comprising only a handful of possible states, and simple agents which learned independently about each one, a so-called \textit{tabular state representation}. By design, this kind of representation fails to support generalization -- the ability to apply what is learned about one state to other similar states -- a shortcoming that becomes increasingly inefficient as environments become larger and more complex, and individual states are therefore less likely to recur.  

One important approach to attaining generalization across states is referred to as \textit{function approximation} \citep{sutton2018reinforcement}, which attempts to assign similar representations to states in which similar actions are required. In one simple implementation of this approach, called \textit{linear} function approximation, each state or situation is encoded as a set of features, and the learner uses a linear readout of these as a basis for selecting its actions.

Although linear function approximation has been often employed in RL research, it has long been recognized that what is needed for RL to produce intelligent, human-like behavior is some form of \textit{non-linear} function approximation. Just as recognizing visual categories (e.g., `cat') is well known to require non-linear processing of visual features (edges, textures, and more complex configurations), non-linear processing of perceptual inputs is generally required in order to decide on adaptive actions. 
 
In acknowledgement of this point, RL research has long sought workable methods for non-linear function approximation. Although a variety of approaches have been explored over the years -- often treating the representation learning problem independent of the underlying RL problem \citep{mahadevan2007proto,konidaris2011value} -- a longstanding aspiration has been to perform adaptive nonlinear function approximation using deep neural networks. 

\subsection*{Deep learning}

\begin{figure}[ht]
    \centering
    \includegraphics[width=1.0\textwidth]{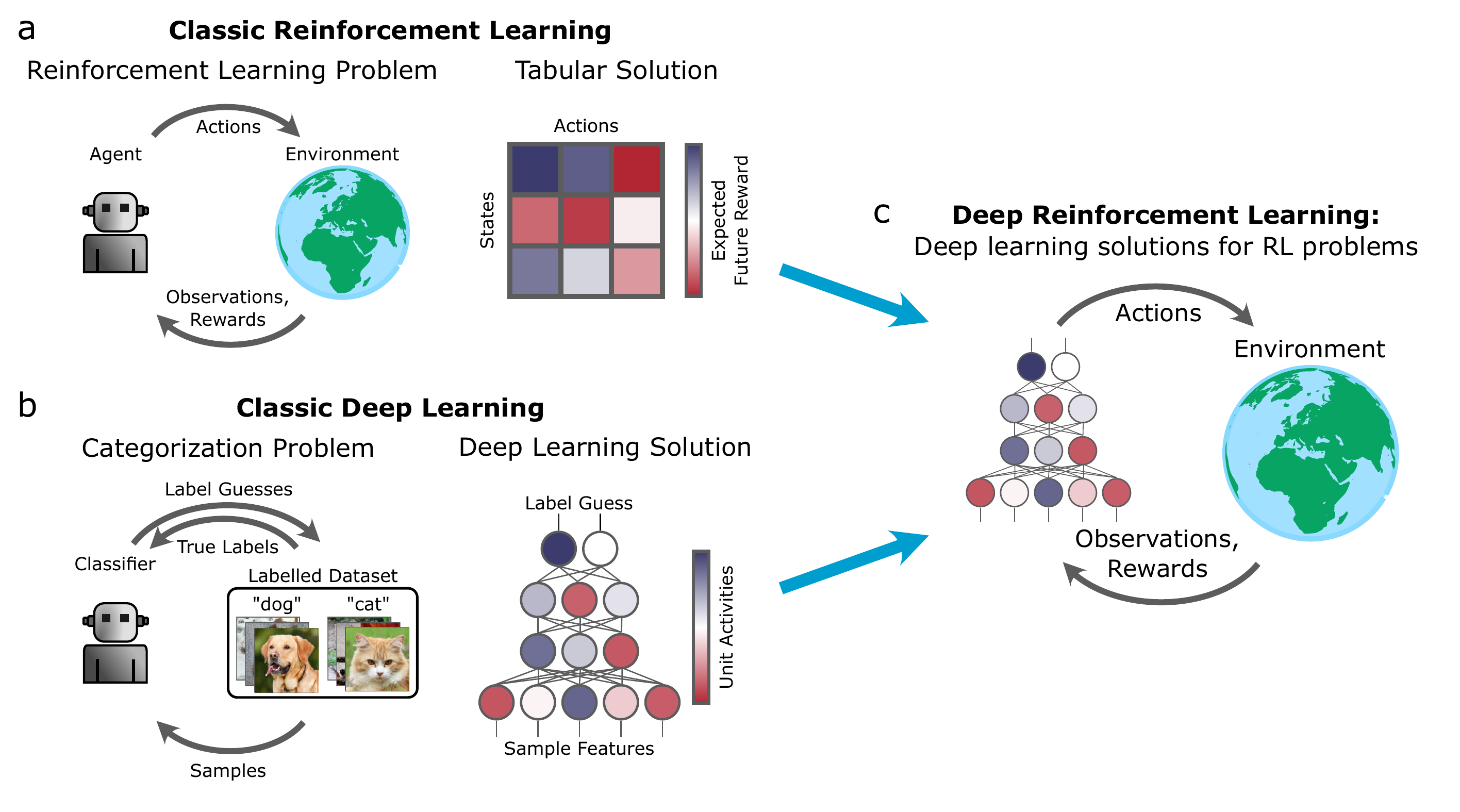}
\caption{ \textbf{RL, deep learning, and deep RL.} 
\textbf{a} 
\textit{Left} The reinforcement learning problem. The agent selects actions and transmits them to the environment, which in turn transmits back to the agent observations and rewards. The agent attempts to select the actions which will maximize long-term reward. The best action might not result in immediate reward, but might instead change the state of the environment to one in which reward can be obtained later. 
\textit{Right} Tabular solution to a reinforcement learning problem. The agent considers the environment to be in one of several discrete states, and learns from experience the expected long-term reward associated with taking each action in each state. These reward expectations are learned independently, and do not generalize to new states or new actions. 
\textbf{b}
\textit{Left} The supervised learning problem. The agent receives a series of unlabeled data samples (e.g. images), and must guess the correct labels. Feedback on the correct label is provided immediately.
\textit{Right} Deep learning solution to a supervised learning problem. The features of a sample (e.g. pixel intensities) are passed through several layers of artificial neurons (circles). The activity of each neuron is a weighted sum of its inputs, and its output is a nonlinear function of its activity. The output of the network is translated into a guess at the correct label for that sample. During learning, network weights are tuned such that these guesses come to approximate the true labels. These solutions have been found to generalize well to samples on which they have not been trained.
\textbf{c}, Deep reinforcement learning, in which a neural network is used as an agent to solve a reinforcement learning problem. By learning appropriate internal representations, these solutions have been found to generalize well to new states and actions. 
}
\label{fig:deep-rl}
\end{figure}

 Deep neural networks are computational systems composed of neuron-like units connected through synapse-like contacts (Figure~\ref{fig:deep-rl}b). Each unit transmits a scalar value, analogous to a spike rate, which is computed based on the sum of its inputs, that is, the activities of ‘upstream’ units multiplied by the strength of the transmitting synapse or connection \citep{goodfellow2016deep}. Critically, unit activity is a non-linear function of these inputs, allowing networks with layers of units interposed between the ‘input’ and ‘output’ sides of the system – i.e., ‘deep’ neural networks -- to approximate any function mapping activation inputs to activation outputs \citep{sutskever2008deep}. Furthermore, when the connectivity pattern includes loops, as in ‘recurrent’ neural networks, the network's activations can preserve information about past events, allowing the network to compute functions based on sequences of inputs.

\textit{Deep learning} refers to the problem of adjusting the connection weights in a deep neural network so as to establish a desired input-output mapping. Although a number of algorithms exist for solving this problem, by far the most efficient and widely used is backpropagation, which uses the chain rule from calculus to decide how to adjust weights throughout a network.

Although backpropagation was developed well over thirty years ago \citep{rumelhart1985learning, Werbos:74}, until recently it was employed almost exclusively for supervised learning, as defined above, or for unsupervised learning, where only inputs are presented, and the task is to learn a ‘good’ representation of those inputs based on some function evaluating representational structure, as is done for example in clustering algorithms. Importantly, both of these learning problems differ fundamentally from RL. In particular, unlike supervised and unsupervised learning, RL requires exploration, since the learner is responsible for discovering actions that increase reward. Furthermore, exploration must be balanced against leveraging action-value information already acquired, or as it is conventionally put, exploration must be weighed against ‘exploitation.’ Unlike with most traditional supervised and unsupervised learning problems, a standard assumption in RL is that the actions of the learning system affect its inputs on the next time-step, creating a sensory-motor feedback loop, and potential difficulties due to nonstationarity in the training data. This creates a situation in which target behaviors or outputs involve multi-step decision processes, rather than single input-output mappings. Until very recently, applying deep learning to RL settings has stood as a frustratingly impenetrable problem.
 
\subsection*{Deep reinforcement learning}

Deep RL leverages the representational power of deep learning to tackle the RL problem. We define a deep RL system as any system that solves an RL problem (i.e. maximizes long-term reward), using representations that are themselves learned by a deep neural network (rather than stipulated by the designer). Typical,  deep RL systems use a deep neural network to compute a non-linear mapping from perceptual inputs to action-values \citep[e.g.,][]{mnih2015human} or action-probabilities \citep[e.g.,][]{silver2016mastering}), as well as reinforcement learning signals that update the weights in this network, often via backpropagation, in order to produce better estimates of reward or to increase the frequency of highly-rewarded actions (Figure~\ref{fig:deep-rl}c).
 
A notable early precursor to modern-day successes with deep RL occurred in the early 1990’s, with a system nicknamed ‘TD-Gammon,’ which combined neural networks with RL to learn how to play backgammon competitively with top human players \citep{tesauro1994td}. More specifically, TD-Gammon used a temporal difference RL algorithm, which computed an estimate for each encountered board position of how likely the system was to win (a \textit{state-value estimate}). The system then computed a \textit{reward-prediction error} (RPE) – essentially an indication of positive surprise or disappointment --  based on subsequent events. The RPE was fed as an error signal into the backpropagation algorithm, which updated the network’s weights so as to yield more accurate state value estimates. Actions could then be selected so as to maximize the state value for the next board state. In order to generate many games on which to train, TD-gammon used self-play, in which the algorithm would play moves against itself until one side won.
 
Although TD-Gammon provided a tantalizing example of what RL implemented via neural networks might deliver, its approach yielded disappointing results in other problem domains. The main issue was instability; whereas in tabular and linear systems, RL reliably moved toward better and better behaviors, when combined with neural networks, the models often collapsed or plateaued, yielding poor results.

This state of affairs changed dramatically in 2013, with the report of the Deep Q Network (DQN), the first deep RL system that learned to play classic Atari video games \citep{mnih2013playing,mnih2015human}. Although DQN was widely noted for attaining better-than-human performance on many games, the real breakthrough was simply in getting deep RL to work in a reliably stable way. It incorporated several mechanisms that reduced nonstationarity, treating the RL problem more like a series of supervised learning problems, upon which the tools of deep learning could be more reliably applied. One example is `experience replay' \citep{lin1991programming}, in which past state-action-reward-next-state transitions were stored away and intermittently re-presented in random order in order to mimic the random sampling of training examples that occurs in supervised learning. This helped to greatly reduce variance and stabilize the updates.

Following DQN, work on deep RL has progressed and expanded at a remarkable pace. Deep RL has been scaled up to highly complex game domains ranging from Dota \citep{berner2019dota} to StarCraft II \citep{vinyals2019grandmaster} to capture-the-flag \citep{jaderberg2019human}. Novel architectures have been developed that support effective deep RL in tasks requiring detailed long-term memory \citep{graves2016hybrid,wayne2018unsupervised}. Deep RL has been integrated with model-based planning, resulting in super-human play in complex games including chess and go \citep{silver2016mastering,silver2017mastering,silver2017mastering2,silver2018general}. Further, methods have been developed to allow deep RL to tackle difficult problems in continuous motor control, including simulations of soccer and gymnastics \citep{merel2018hierarchical,heess2016learning}, and robotics problems such as in-hand manipulation of a Rubik's cube \cite{akkaya2019solving}. We review some of these developments in greater detail below, as part of a larger consideration of what implications deep RL may have for neuroscience, the topic to which we now turn.
 
\section*{Deep RL and Neuroscience}

\begin{figure}[h!]
    \centering
    \includegraphics[width=1.0\textwidth]{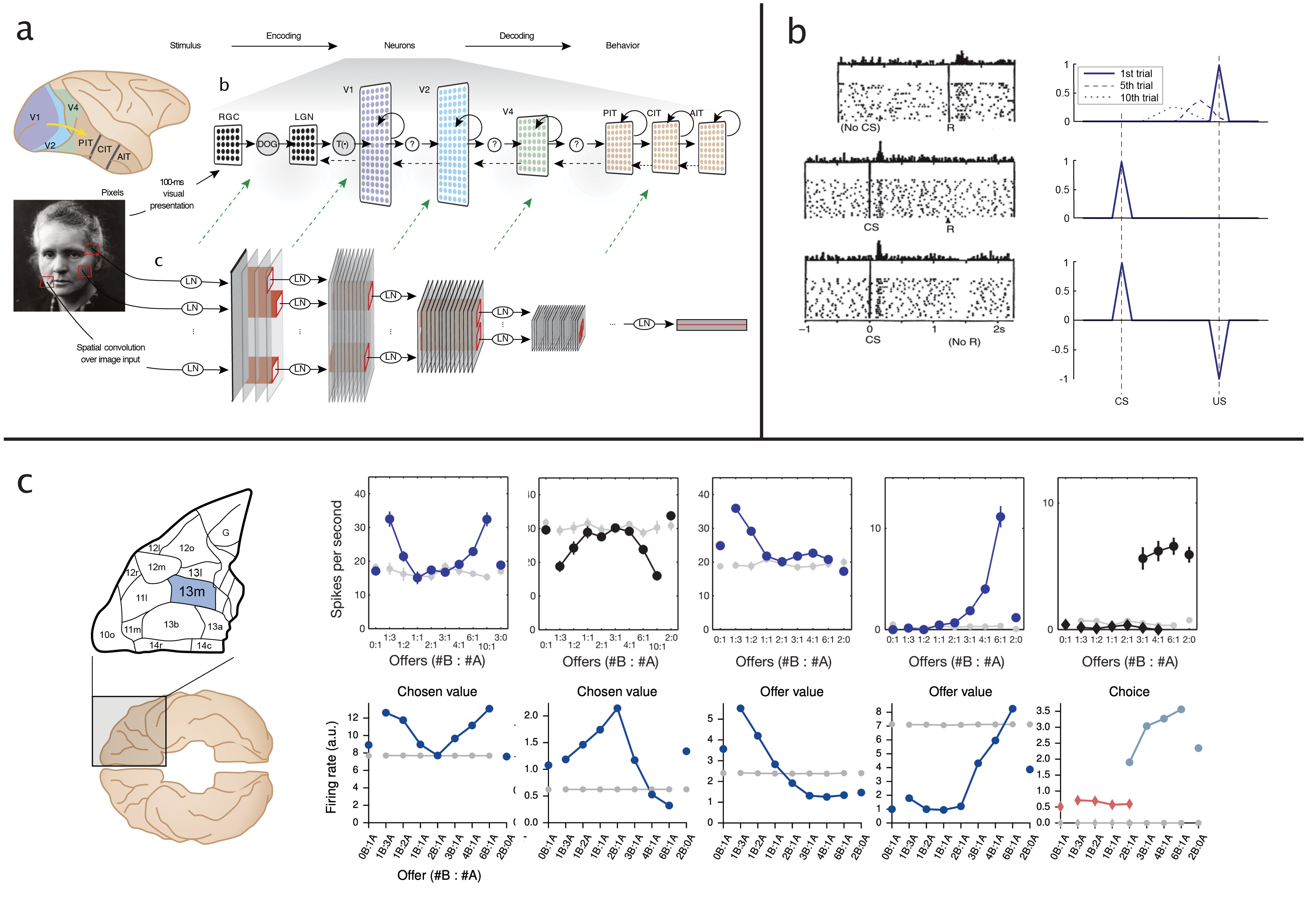}
\caption{ \textbf{Applications to neuroscience.} 
\textbf{a.} Supervised deep learning has been used in a wide range of studies to model and explain neural activity. In one influential study, \citeauthor{yamins2016using} \citep{yamins2016using} employed a deep convolutional network (shown schematically in the lower portion of the figure) to model single-unit responses in various portions of the macaque ventral stream (upper portion). Figure adapted from \citep{yamins2016using}. 
\textbf{b.} Reinforcement learning has been connected with neural function in a number of ways. Perhaps most impactful has been the link established between phasic dopamine release and the temporal-difference reward-prediction error signal or RPE. The left side of the figure panel shows typical spike rasters and histograms from dopamine neurons in ventral tegmental area under conditions where a food reward arrives unpredictably (top), arrives following a predictive cue (CS), or is withheld following a CS. The corresponding panels on the right plots RPEs from a temporal-difference RL model under parallel conditions, showing qualitatively identical dynamics. Figure adapted from \citep{niv2009reinforcement}.
\textbf{c.} Applications of deep RL to neuroscience have only just begun. In one pioneering study \citeauthor{song2017reward} \citep{song2017reward} trained a recurrent deep RL network on a reward-based decision making task paralleling one that had been studied in monkeys by \citeauthor{padoa2006neurons} \citep{padoa2006neurons}. The latter study examined the responses of neurons in orbitofrontal area 13m (see left panel) across many different choice sets involving two flavors of juice in particular quantities (x-axes in upper plots), reporting neurons whose activity tracked the inferred value of the monkey's preferred choice (two top left panels), the value of each individual juice (next two panels), or the identity of the juice actually chosen (right panel). Examining units within their deep RL model, \citeauthor{song2017reward} found patterns of activity closely resembling the neurophysiological data (bottom panels). Panels adapted from \citep{song2017reward}, \citep{padoa2006neurons} and \citep{stalnaker2015orbitofrontal}.
}
\label{fig:neuroExamples}
\end{figure}

Deep RL is built from components – deep learning and RL – that have already independently had a profound impact within neuroscience. Deep neural networks have proven to be an outstanding model of neural representation \citep{yamins2014performance,sussillo2015neural,kriegeskorte2015deep,mante2013context, pandarinath2018inferring, rajan2016recurrent,zipser1991recurrent,zipser1988back} (Figure~\ref{fig:neuroExamples}a). However, this research has for the most part utilized supervised training, and has therefore provided little direct leverage on the big-picture problem of understanding motivated, goal-directed behavior within a sensory-motor loop. At the same time, reinforcement learning has provided a powerful theory of the neural mechanisms of learning and decision making \citep{niv2009reinforcement}. This theory most famously explains activity of dopamine neurons as a reward prediction error  \citep{watabe2017neural,glimcher2011understanding,lee2012neural,daw2014multiple} (Figure~\ref{fig:neuroExamples}b), but also accounts for the role of a wide range of brain structures in reward-driven learning and decision making \citep{stachenfeld2017,botvinick2009hierarchically,o2006making,glascher2010states,wang2018prefrontal,wilson2014orbitofrontal}. It has been integrated into small neural networks with handcrafted structure to provide models of how multiple brain regions may interact to guide learning and decision-making \citep{o2006making,frank2006decision}. Just as in the machine learning context, however, RL itself has until recently offered neuroscience little guidance in thinking about the problem of representation \citep[for discussion, see][]{botvinick2015reinforcement,wilson2014orbitofrontal,stachenfeld2017,behrens2018cognitive, gershman2010context}.

Deep RL offers neuroscience something new, by showing how RL and deep learning can fit together. While deep learning focuses on how representations are learned, and RL on how rewards guide learning, in deep RL new phenomena emerge: processes by which representations support, and are shaped by, reward-driven learning and decision making. 

If deep RL offered no more than a concatenation of deep learning and RL in their familiar forms, it would be of limited import. But deep RL is more than this; when deep learning and RL are integrated, each triggers new patterns of behavior in the other, leading to computational phenomena unseen in either deep learning or RL on their own. That is to say, deep RL is much more than the sum of its parts. And the novel aspects of the integrated framework in turn translate into new explanatory principles, hypotheses and available models for neuroscience.

We unpack this point in the next section in considering some of the few neuroscience studies to have appeared so far that have leveraged deep RL, turning subsequently to a consideration of some wider issues that deep RL raises for neuroscience research.
 
\section*{Vanguard studies}

Although a number of commentaries have appeared which address aspects of deep RL from a neuroscientific perspective \citep{hassabis2017neuroscience,zador2019critique,marblestone2016toward}, few studies have yet applied deep RL models directly to neuroscientific data.

In a few cases, researchers have deployed deep RL in ways analogous to previous applications of deep learning and RL. For example, transplanting a longstanding research strategy from deep learning \citep{yamins2014performance,zipser1991recurrent} to deep RL, \cite{song2017reward} trained a recurrent deep RL model on a series of reward-based decision making tasks that have been studied in the neuroscience literature, reporting close correspondences between the activation patterns observed in the network’s internal units and neurons in dorsolateral prefrontal, orbitofrontal and parietal cortices (Figure~\ref{fig:neuroExamples}c). Work by \cite{banino2018vector} combined supervised deep learning and deep RL methods to show how grid-like representations resembling those seen in entorhinal cortex can enhance goal-directed navigation performance.

\begin{figure}[ht]
    \centering
    \includegraphics[width=1.0\textwidth]{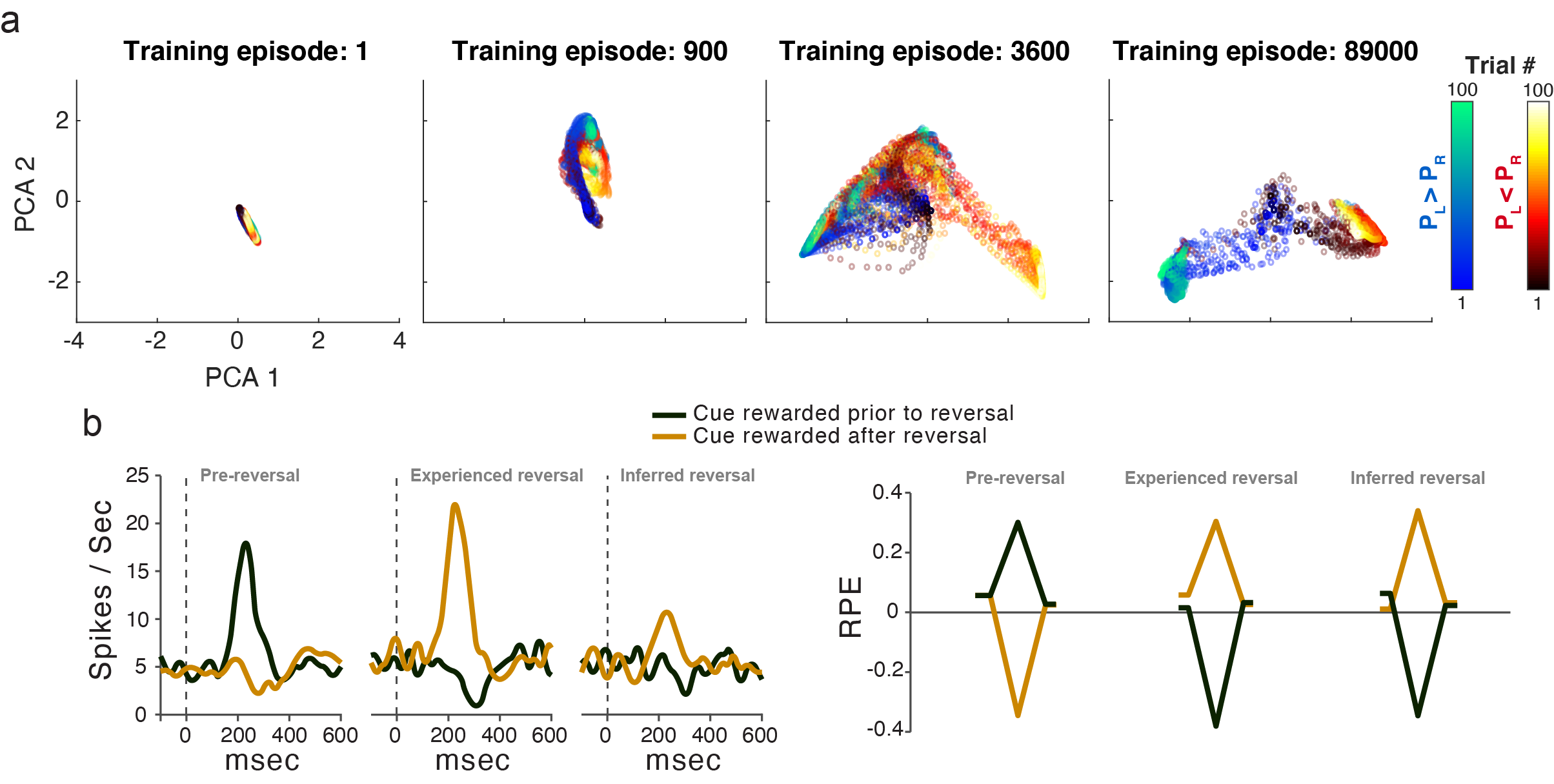}
\caption{ \textbf{Meta-reinforcement learning.}  \textbf{a,} Visualization of representations learned through meta-reinforcement learning, at various stages of training. An artificial agent is trained on a series of independent Bernoulli 2-armed bandits (100 trials per episode), such that the probability of reward payout $P_L$ and $P_R$ are drawn uniformly from $\mathcal{U}$(0, 1). Scatter points depict the first two principal components of the RNN activation (LSTM output) vector taken from evaluation episodes at certain points in training, colored according to trial number (darker = earlier trials) and whether $P_L > P_R$. Only episodes for which $|P_L - P_R| > 0.3$ are plotted. \textbf{b,} Panels adapted from \citep{bromberg2010pallidus} and \citep{wang2018prefrontal}. \textit{left}, Dopaminergic activity in response to cues ahead of a reversal and for cues with an experienced and inferred change in value. \textit{right}, Corresponding RPE signals from an artificial agent. Leading and trailing points for each data-series correspond to initial fixation and saccade steps. Peaks/troughs correspond to stimulus presentation.}
\label{fig:metarl}
\end{figure}

As we have stressed, phenomena arise within deep RL that do not arise in deep learning or RL considered separately. A pair of recent studies have focused on the neuroscientific implications of these emergent phenomena. In one, \cite{wang2018prefrontal} examined the behavior of recurrent deep RL systems, and described a novel \textit{meta-reinforcement learning} effect: When trained on a series of interrelated tasks – for example, a series of forced-choice decision tasks with the same overall structure but different reward probabilities -- recurrent deep RL networks develop the ability to adapt to new tasks of the same kind without weight changes. This is accompanied by correspondingly structured representations in the activity dynamics of the hidden units that emerge throughout training (Figure~\ref{fig:metarl}a). Slow RL-driven learning at the level of the network’s connection weights shape the network’s activation dynamics such that rapid behavioral adaptation can be driven by those activation dynamics alone, akin to the idea from neuroscience that RL can be supported, in some cases, by activity-based working memory \citep{collins2012much}. In short, slow RL spontaneously gives rise to a separate and faster RL algorithm. Wang and colleagues showed how this meta-reinforcement learning effect could be applied to explain a wide range of previously puzzling findings from neuroscientific studies of dopamine and prefrontal cortex function (Figure~\ref{fig:metarl}b). 

\begin{figure}[h!]
    \centering
    \includegraphics[width=0.75\textwidth]{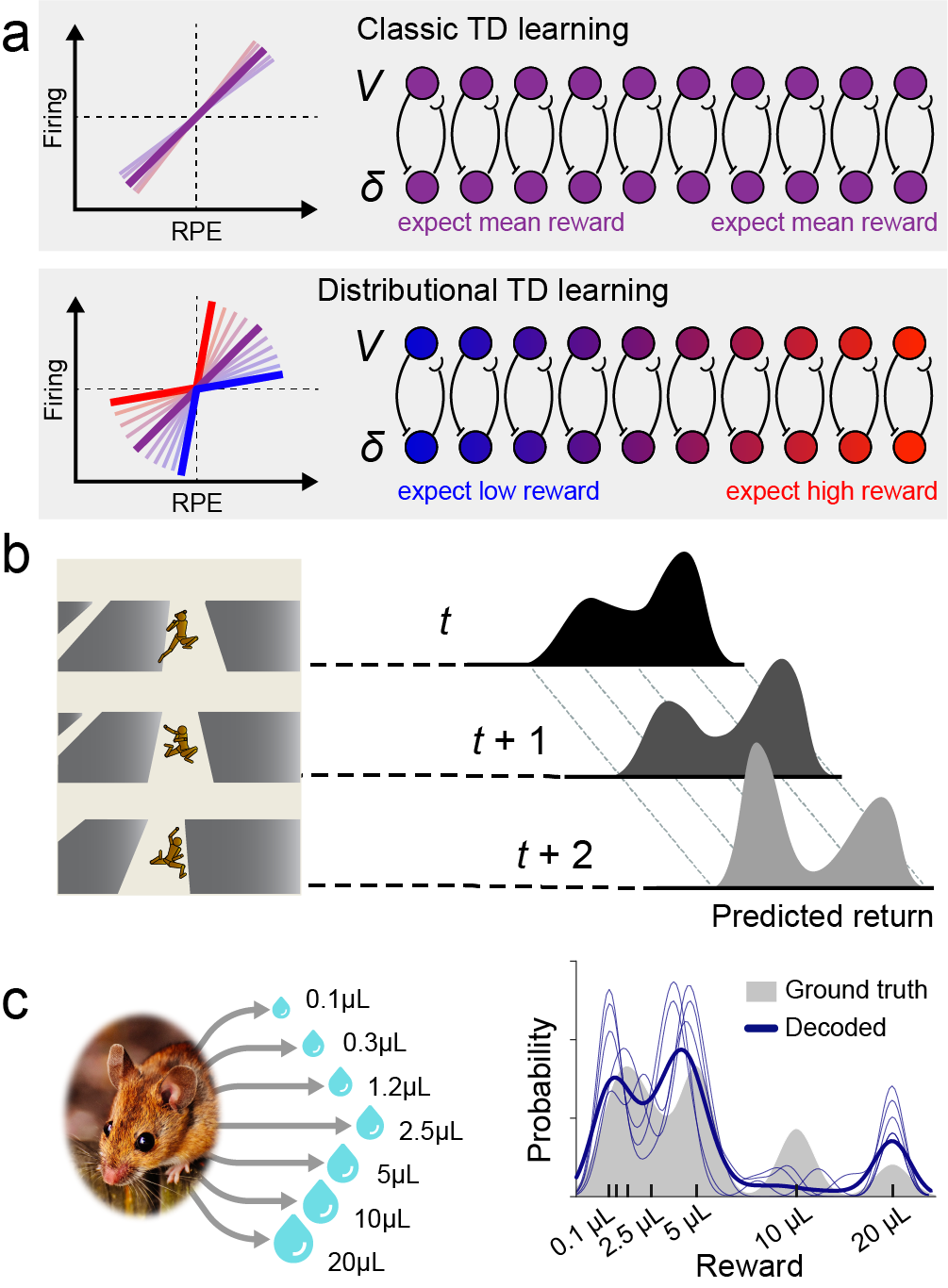}
\caption{ \textbf{Distributional RL.} 
\textbf{a,} \textit{top}, In the classic temporal difference (TD) model, each dopamine cell computes a prediction error with respect to the same predicted reward. \textit{bottom}, In distributional TD, some RPE channels amplify negative RPEs (blue) and others amplify positive RPEs (red). This causes the channels to learn different reward predictions, ranging from very pessimistic (blue) to very optimistic (red). 
\textbf{b,} Artificial agents endowed with diverse RPE scaling learn to predict the return distribution. In this example, the agent is uncertain whether it will successfully land on the platform. The agent's predicted reward distribution on three consecutive timesteps is shown at right.
\textbf{c,} In real animals, it is possible to decode the reward distribution directly from dopamine activity. Here, mice were extensively trained on a task with probabilistic reward. The actual reward distribution of the task is shown as a gray shaded area. When interpreted as RPE channels of a distributional TD learner, the firing of dopamine cells decodes to the distribution shown in blue (thin traces are best five solutions and thick trace is their mean). The decoded distribution matches multiple modes of the actual reward distribution. Panels adapted from \citep{dabney2020distributional}.}
\label{fig:distrl}
\end{figure}

A second such study comes from \cite{dabney2020distributional}. This leveraged a deep RL technique developed in recent AI work, and referred to as \textit{distributional RL} \citep{bellemare2017distributional}. Earlier, in discussing the history of deep RL, we mentioned the reward-prediction error or RPE. In conventional RL, this signal is a simple scalar, with positive numbers indicating a positive surprise and negative ones indicating disappointment. More recent neuroscientifically inspired models have suggested that accounting for the distribution and uncertainty of reward is important for decision-making under risk \citep{mikhael2016learning}. In distributional RL, the RPE is expanded to a vector, with different elements signaling RPE signals based on different \textit{a priori} forecasts, ranging from highly optimistic to highly pessimistic predictions (Figure~\ref{fig:distrl}a,b).  This modification had been observed in AI work to dramatically enhance both the pace and outcome of RL across a variety of tasks, something – importantly – which is observed in deep RL, but not simpler forms such as tabular or linear RL (due in part to the impact of distributional coding on representation learning \citep{lyle2019comparative}). Carrying this finding into the domain of neuroscience, Dabney and colleagues studied electrophysiological data from mice to test whether the dopamine system might employ the kind of vector code involved in distributional RL. As noted earlier, dopamine has been proposed to transmit an RPE-like signal. Dabney and colleagues obtained strong evidence that this dopaminergic signal is distributional, conveying a spectrum of RPE signals ranging from pessimistic to optimistic (Figure~\ref{fig:distrl}c).

\section*{Topics for next-step research}
 
As we have noted, explorations of deep RL in neuroscience have only just begun. What are the key opportunities going forward? In the sections below we outline six areas where it appears deep RL may provide leverage for neuroscientific research. In each case, intensive explorations are already underway in the AI context, providing neuroscience with concrete opportunities for translational research. While we stress tangible proposals in what follows, it is important to bear in mind that these proposals do not restrict the definition of deep RL. Deep RL is instead a broad and multi-faceted framework, within which algorithmic details can be realized in a huge number of ways, making the space of resulting hypotheses for neuroscience bracingly diverse.
 
\subsection*{Representation learning}
 
\begin{figure}[h!]
    \centering
    \includegraphics[width=1.0\textwidth]{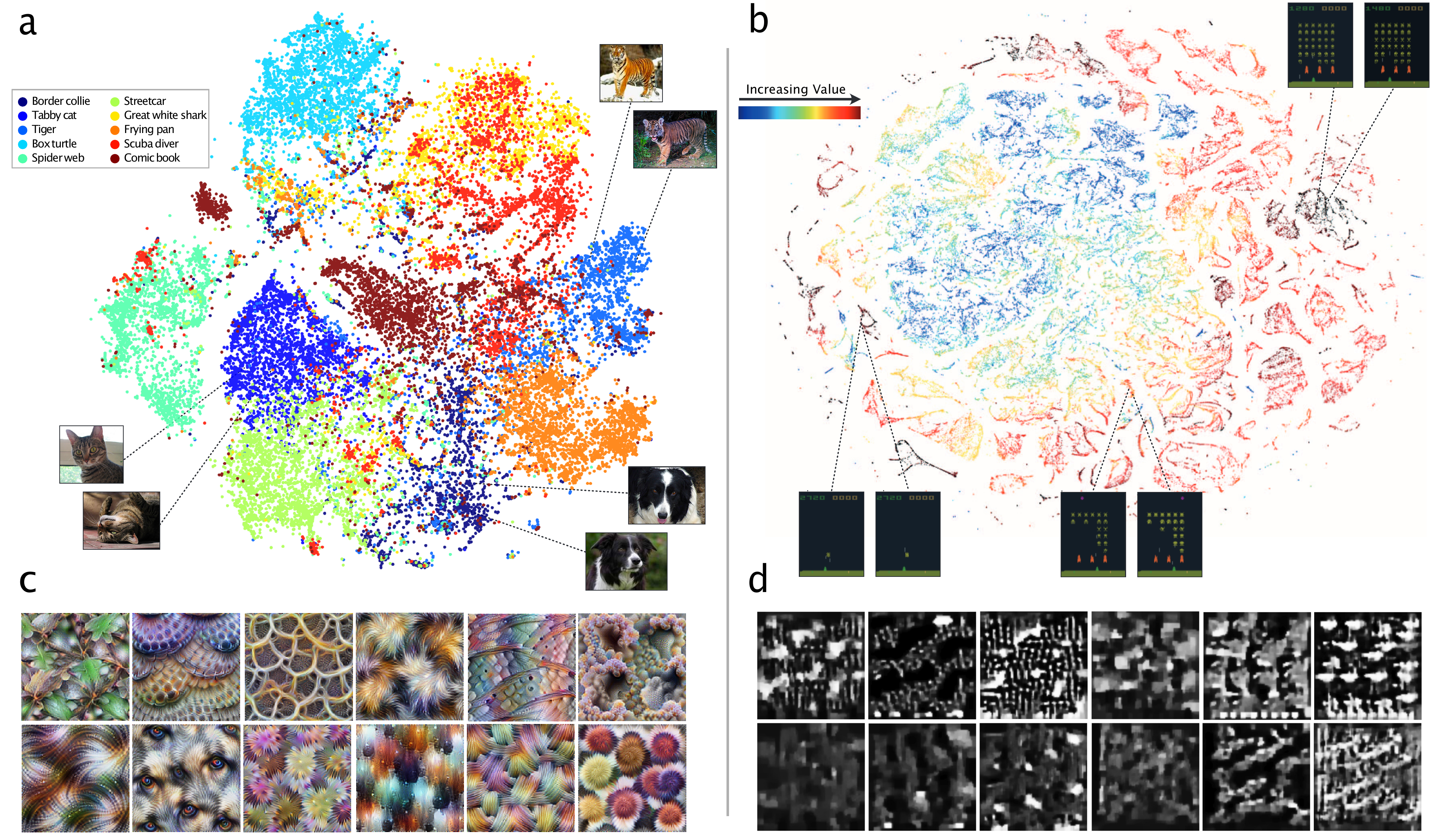}
\caption{ \textbf{Representations learned by deep supervised learning and deep RL.} \textbf{a,} Representations of natural images (ImageNet \citep{deng2009imagenet}) from a deep neural network trained to classify objects \citep{carter2019activation}. The t-SNE embedding of representations in one layer (``mixed5b''), coloured by predicted object class, and with example images shown. \textbf{b,} Synthesized inputs that maximally activate individual artificial neurons (in layer ``mixed4a'') show specialization for high-level features and textures to support object recognition \citep{olah2017feature}. \textbf{c,} Representations of Atari video game images \citep{bellemare2013arcade}, from a DQN agent trained with deep RL \citep{mnih2015human}. The t-SNE embedding of representations from the final hidden layer, coloured by predicted future reward value, and with example images shown. \textbf{d,} Synthesized images that maximally activate individual cells from the final convolutional layer reveal texture-like detail for reward-predictive features \citep{such2019atari}. For example, in the game Seaquest, the relative position of the Submarine to incoming fish appears to be captured in the top-rightmost image.}
\label{fig:tsne}
\end{figure}

The question of representation has long been central to neuroscience, beginning perhaps with the work of Hubel and Weisel \citep{hubel1959receptive} and continuing robustly to the present day \citep{constantinescu2016organizing, stachenfeld2017, wilson2014orbitofrontal}. Neuroscientific studies of representation have benefited from tools made available by deep learning \citep{zipser1988back,yamins2014performance}, which provides models of how representations can be shaped by sensory experience. Deep RL expands this toolkit, providing for the first time models of how representations can be shaped by rewards and by task demands. In a deep RL agent, reward-based learning shapes internal representations, and these representations in turn support reward-based decision making. A canonical example would be the DQN network training on an Atari task. Here, reward signals generated based on how many points are scored feed into a backpropagation algorithm that modifies weights throughout the deep neural network, updating the response profiles of all units. This results in representations that are appropriate for the task. Whereas a supervised learning system assigns similar representations to images with similar labels (Figure ~\ref{fig:tsne}a,b), deep RL tends to associate images with similar functional task implications (Figure ~\ref{fig:tsne}c,d). 
 
This idea of reward-based representation learning resonates with a great deal of evidence from neuroscience. We know, for example, that representations of visual stimuli in prefrontal cortex depend on which task an animal has been trained to perform \citep{freedman2001categorical}, and that effects of task reward on neural responses can be seen even in primary visual cortex \citep{pakan2018action}.
 
The development and use of deep RL systems has raised awareness of two serious drawbacks of representations that are shaped by RL alone. One problem is that task-linked rewards are generally sparse. In chess, for example, reward occurs once per game, making it a weak signal for learning about opening moves. A second problem is over-fitting -- internal representations shaped exclusively by task-specific rewards may end up being useful only for tasks the learner has performed, but completely wrong for new tasks \citep{zhang2018study,cobbe2019quantifying}. Better would be some learning procedure that gives rise to internal representations that are more broadly useful, supporting transfer between tasks.

To address these issues, deep reinforcement learning is often supplemented in practice with either unsupervised learning \citep{higgins2017darla}, or ‘self-supervised’ learning. In self-supervised learning the agent is trained to produce, in addition to an action, some auxiliary output which matches a training signal that is naturally available from the agent’s stream of experience, regardless of what specific RL task it is being trained on \citep{jaderberg2016reinforcement, banino2018vector}. An example is prediction learning, where the agent is trained to predict, based on its current situation, what it will observe on future timesteps \citep{wayne2018unsupervised,gelada2019deepmdp}. Unsupervised and self-supervised learning mitigate both problems associated with pure reinforcement learning, since they shape representations in a way that is not tied exclusively to the specific tasks confronted by the learner, thus yielding representations that have the potential to support transfer to other tasks when they arise. All of this is consistent with existing work in neuroscience, where unsupervised learning \citep[e.g.,][]{olshausen1996emergence,hebb1949organization,kohonen2012self} and prediction learning \citep[e.g.,][]{schapiro2013neural,stachenfeld2017,rao1999predictive} have been proposed to shape internal representations. Deep RL offers the opportunity to pursue these ideas in a setting where these forms of learning can mix with reward-driven learning \citep{marblestone2016toward,richards2019deep} and where the representations they produce support adaptive behavior.
 
One further issue foregrounded in deep RL involves the role of inductive biases in shaping representation learning. Most deep RL systems that take visual inputs employ a processing architecture \citep[a convolutional network,][]{fukushima1980neocognitron} that biases them toward representations that take into account the translational invariance of images. And more recently developed architectures build in a bias to represent visual inputs as comprising sets of discrete objects with recurring pairwise relationships \citep{watters2019cobra,battaglia2018relational}. Such ideas recall existing neuroscientific findings \citep{roelfsema1998object}, and have interesting consequences in deep RL, such as the possibility of exploring and learning much more efficiently by decomposing the environment into objects \citep{diuk2008object,watters2019cobra}.

\subsection*{Model-based RL}
 
An important classification of reinforcement learning algorithms is between `model-free' algorithms, which learn a direct mapping from perceptual inputs to action outputs, and `model-based' algorithms, which instead learn a ‘model’ of action-outcome relationships and use this to plan actions by forecasting their outcomes.

This dichotomy has had a marked impact in neuroscience, where brain regions have been accorded different roles in these two kinds of learning, and where an influential line of research has focused on how the two forms of learning may trade off against one another \citep{lee2014neural,daw2011model,balleine1998goal,daw2005uncertainty,dolan2013goals}. Deep RL opens up a new vantage point on the relationship between model-free and model-based RL. For example, in AlphaGo and its successor systems \cite{silver2016mastering,silver2017mastering,silver2018general}, model-based planning is guided in part by value estimates and action tendencies learned through model-free RL. Related interactions between the two systems have been studied in neuroscience and psychology \citep{cushman2015habitual,keramati2016adaptive}.

In AlphaGo, the action-outcome model used in planning is hand-coded. Still more interesting, from a neuroscientific point of view \citep{glascher2010states}, is recent work in which model-based RL relies on models learned from experience \cite{schrittwieser2019mastering,nagabandi2018neural,ha2018world}. Although these algorithms have achieved great success in some domains, a key open question is whether systems can learn to capture transition dynamics at a high-level of abstraction (“If I throw a rock at that window, it will shatter”) rather than being tied to detailed predictions about perceptual observations (predicting where each shard would fall) \citep{behrens2018cognitive, konidaris2019necessity}.
 
One particularly intriguing finding from deep RL is that there are circumstances under which processes resembling model-based RL may emerge spontaneously within systems trained using model-free RL algorithms \citep{wang2016learning,guez2019investigation}. The neuroscientific implications of this `model-free planning' have already been studied in a preliminary way  \citep{wang2018prefrontal}, but it deserves further investigation. Intriguingly, model-based behavior is also seen in RL systems that employ a particular form of predictive code, referred to as the ‘successor representation’ \citep{vertes2019neurally,momennejad2020learning,kulkarni2016deep,barreto2017successor}, suggesting one possible mechanism through which model-free planning might arise. 

An interesting question that has arisen in neuroscientific work is how the balance between model-free and model-based RL is arbitrated, that is, what are the mechanisms that decide, moment to moment, whether behavior is controlled by model-free or model-based mechanisms \citep{daw2005uncertainty,lee2014neural}. Related to this question, some deep RL work in AI has introduced mechanisms that learn through RL whether and how deeply to plan before committing to an action \citep{hamrick2017metacontrol}. The resulting architecture is reminiscent of work from neuroscience on cognitive control mechanisms implemented in the prefrontal cortex \citep{botvinick2014computational}, a topic we discuss further below.

\subsection*{Memory}
 
On the topic of memory, arguably one of the most important in neuroscience, deep RL once again opens up fascinating new questions and highlights novel computational possibilities. In particular, deep RL provides a computational setting in which to investigate how memory can support reward-based learning and decision making, a topic which has been of growing interest in neuroscience \citep[see, e.g.,][]{eichenbaum1999hippocampus,gershman2017reinforcement}. 
The first broadly successful deep RL models relied on experience replay \citep{mnih2013playing}, wherein past experiences are stored, and intermittently used alongside new experiences to drive learning. This has an intriguing similarity to the replay events observed in hippocampus and elsewhere, and indeed was inspired by this phenomenon and its suspected role in memory consolidation \citep{wilson1994reactivation,kumaran2016learning}. While early deep RL systems replayed experience uniformly, replay in the brain is not uniform \citep{mattar2018prioritized,gershman2017reinforcement,gupta2010hippocampal, carey2019reward}, and non-uniformity has been explored in machine learning as a way to enhance learning \citep{schaul2015prioritized}.

In addition to driving consolidation, memory maintenance and retrieval in the brain is also used for online decision-making \citep{pfeiffer2013hippocampal,wimmer2012preference,bornstein2017reinstated,o2006making}. In deep RL, two kinds of memory serve this function. First, `episodic' memory systems read and write long-term storage slots \citep{wayne2018unsupervised, lengyel2008hippocampal, blundell2016model}. One interesting aspect of these systems is that they allow relatively easy analysis of what information is being stored and retrieved at each timestep \citep{graves2016hybrid,Banino2020MEMO}, inviting comparisons to neural data. Second, recurrent neural networks store information in activations, in a manner similar to what is referred to in neuroscience as working memory maintenance. The widely used `LSTM' and `GRU' architectures use learnable gating to forget or retain task-relevant information, reminiscent of similar mechanisms which have been proposed to exist in the brain \citep{chatham2015multiple,stalter2020dopamine}.

Still further deep-RL memory mechanisms are being invented at a rapid rate, including systems that deploy attention and relational processing over information in memory \citep[e.g.,][]{parisotto2019stabilizing,graves2016hybrid} and systems that combine and coordinate working and episodic memory \citep[e.g.,][]{ritter2018been}. This represents one of the topic areas where an exchange between deep RL and neuroscience seems most actionable and most promising.

\subsection*{Exploration}
 
As noted earlier, exploration is one of the features that differentiates RL from other standard learning problems. RL imposes the need to seek information actively, testing out novel behaviors and balancing them against established knowledge, negotiating the explore-exploit trade-off.  Animals, of course, face this challenge as well, and it has been of considerable interest in neuroscience and psychology \citep[see e.g.,][]{costa2019subcortical,gershman2018deconstructing,wilson2014humans,schwartenbeck2013exploration}. Here once again, deep RL offers a new computational perspective and a set of specific algorithmic ideas.
 
A key strategy in work on exploration in RL has been to include an auxiliary (\textit{`intrinsic'}) reward \citep{schmidhuber1991curious,dayan2002reward,chentanez2005intrinsically,oudeyer2007intrinsic}, such as for novelty, which encourages the agent to visit unfamiliar states or situations. However, since deep RL generally deals with high-dimensional perceptual observations, it is rare for exactly the same perceptual observation to recur. The question thus arises of how to quantify novelty, and a range of innovative techniques have been proposed to address this problem \citep{bellemare2016unifying,pathak2017curiosity,burda2018exploration,Badia2020Never}. Another approach to intrinsically motivated exploration is to base it not on novelty but on uncertainty, encouraging the agent to enter parts of the environment where its predictions are less confident \citep{osband2016deep}. And still other work has pursued the idea of allowing agents to learn or evolve their own intrinsic motivations, based on task experience \citep{niekum2010genetic,singh2010intrinsically,zheng2018learning}.

Meta-reinforcement learning provides another interesting and novel perspective on exploration. As noted earlier, meta-reinforcement learning gives rise to activation dynamics that support learning, even when weight changes are suspended. Importantly, the learning that occurs in that setting involves exploration, which can be quite efficient because it is structured to fit with the kinds of problems the system was trained on. Indeed, exploration in meta-reinforcement learning systems can look more like hypothesis-driven experimentation than random exploration \citep{denil2016learning,dasgupta2019causal}. These properties of meta-reinforcement learning systems make them an attractive potential tool for investigating the neural basis of strategic exploration in animals.

Finally, some research in deep RL proposes to tackle exploration by sampling randomly in the space of hierarchical behaviors \citep{machado2017laplacian,jinnai2020exploration,Hansen2020Fast}. This induces a form of directed, temporally-extended, random exploration reminiscent of some animal foraging models \citep{viswanathan1999optimizing}.

\subsection*{Cognitive control and action hierarchies}
 
Cognitive neuroscience has long posited a set of functions, collectively referred to as `cognitive control', which guide task selection and strategically organize cognitive activity and behavior \citep{botvinick2014computational}. The very first applications of deep RL contained nothing corresponding to this set of functions. However, as deep RL research has developed, it has begun to grapple with the problem of attaining competence and switching among multiple tasks or skills, and in this context a number of computational techniques have been developed which bear an intriguing relationship with neuroscientific models of cognitive control.
 
Perhaps most relevant is research that has adapted to deep RL ideas originating from the older field of hierarchical reinforcement learning. Here, RL operates at two levels, shaping a choice among high-level multi-step actions (e.g. `make coffee') and also among actions at a more atomic level \citep[e.g. `grind beans'; see][]{botvinick2009hierarchically}. Deep RL research has adopted this hierarchical scheme in a number of ways \citep{bacon2017option,harutyunyan2019termination,barreto2019option,vezhnevets2017feudal}. In some of these, the low-level system can operate autonomously, and the higher-level system intervenes only at a cost which makes up part of the RL objective \citep{teh2017distral,harb2018waiting}, an arrangement that resonates with the notions in neuroscience of habit pathways and automatic versus controlled processing \citep{dolan2013goals,balleine2010human}, as well as the idea of a ‘cost of control’ \citep{shenhav2017toward}.  In deep RL, the notion of top-down control over lower-level habits has also been applied in motor control tasks, in architectures resonating with classical neuroscientific models of hierarchical control \citep{merel2018hierarchical, heess2016learning}.

Intriguingly, hierarchical deep RL systems have in some cases been configured to operate at different time-scales at different levels, with slower updates at higher levels, an organizational principle that resonates with some neuroscientific evidence concerning hierarchically organized time-scales across cortex \citep{badre2008cognitive,hasson2008hierarchy}.

\subsection*{Social cognition}
 
A growing field of neuroscience research investigates the neural underpinnings of social cognition. In the last couple of years deep RL has entered this space, developing methods to train multiple agents in parallel in interesting multi-agent scenarios. This includes competitive team games, where individual agents must learn how to coordinate their actions \citep{jaderberg2019human, berner2019dota}, cooperative games requiring difficult coordination \citep{foerster2019bayesian}, as well as thorny ‘social dilemmas,’ where short-sighted selfish actions must be weighed against cooperative behavior \citep{leibo2017multi}. The behavioral sciences have long studied such situations, and multi-agent deep RL offers new computational leverage on this area of research, up to and including the neural mechanisms underlying mental models of others, or ‘theory of mind’ \citep{rabinowitz2018machine, tacchetti2018relational}.

\section*{Challenges and caveats}
 
It is important to note that deep RL is an active – and indeed quite new -- area of research, and there are many aspects of animal and especially human behavior that it does not yet successfully capture. Arguably, from a neuroscience perspective, these limitations have an upside, in that they throw into relief those cognitive capacities that remain most in need of computational elucidation \citep{lake2017building, zador2019critique}, and indeed point to particular places where neuroscience might be able to benefit AI research.
 
One issue that has already been frequently pointed out is the slowness of learning in deep RL, that is, its demand for large amounts of data. DQN, for example, required much more experience to reach human-level performance in Atari games than would be required by an actual human learner \citep{lake2017building}. This issue is more complicated than it at first sounds, both because standard deep RL algorithms have become progressively more sample efficient, through alternative approaches like meta-learning and deep RL based on episodic memory \citep{ritter2018been, botvinick2019reinforcement}, and because human learners bring to bear a lifetime of prior experiences to each new learning problem.
 
Having said this, it is also important to acknowledge that deep RL systems have not yet been proven to be capable of matching humans when it comes to flexible adaptation based on structured inference, leveraging a powerful store of background knowledge. Whether deep RL systems can close this gap is an open and exciting question. Some recent work suggests that deep RL systems can, under the right circumstances, capitalize on past learning to quickly adapt systematically to new situations that appear quite novel \citep{hill2019emergent}, but this does not invariably happen \citep[see e.g.,][]{lake2017generalization}, and understanding the difference is of interest both to AI and neuroscience.
 
A second set of issues centers on more nuts-and-bolts aspects of how learning occurs. One important challenge, in this regard, is long-term temporal credit assignment, that is, updating behaviour based on rewards that may not accrue until a substantial time after the actions that were responsible for generating them. This remains a challenge for deep RL systems. Novel algorithms have recently been proposed \citep[see for example][]{hung2019optimizing}, but the problem is far from solved, and a dialogue with neuroscience in this area may be beneficial to both fields.
 
More fundamental is the learning algorithm almost universally employed in deep RL research: backpropagation. As has been widely discussed in connection with supervised deep learning research, which also uses backpropagation, there are outstanding questions about how backpropagation might be implemented in biological neural systems, if indeed it is at all \citep{Lillicrap2020NRN,whittington2019theories} \citep[although see][for interesting proposals for how backpropagation might be implemented in biological circuits]{sacramento2018dendritic,payeur2020burst}. And there are inherent difficulties within backpropagation associated with preserving the results of old learning in the face of new learning, a problem for which remedies are being actively researched, in some cases taking inspiration from neuroscience \citep{kirkpatrick2017overcoming}.
 
Finally, while we have stressed alignment of deep RL research with neuroscience, it is also important to highlight an important dimension of mismatch.  The vast majority of contemporary deep RL research is being conducted in an engineering context, rather than as part of an effort to model brain function. As a consequence, many techniques employed in deep RL research are fundamentally unlike anything that could reasonably be implemented in a biological system. At the same time, many concerns that are central in neuroscience, for example energy efficiency or heritability of acquired knowledge across generations, do not arise as natural questions in AI-oriented deep RL research. Of course, even when there are important aspects that differentiate engineering-oriented deep RL systems from biological systems, there may still be high-level insights that can span the divide. Nevertheless, in scoping out the potential for exchange between neuroscience and contemporary deep RL research it is important to keep these potential sources of discrepancy in mind.
 
\section*{Conclusion}
 
The recent explosion of progress in AI offers exciting new opportunities for neuroscience on many fronts. In discussing deep RL, we have focused on one particularly novel area of AI research which, in our view, has particularly rich implications for neuroscience, most of which have not yet been deeply explored. As we have described, deep RL provides an agent-based framework for studying the way that reward shapes representation, and how representation in turn shapes learning and decision making, two issues which together span a large swath of what is most central to neuroscience. We look forward to an increasing engagement in neuroscience with deep RL research. As this occurs there is also a further opportunity. We have focused on how deep RL can help neuroscience, but as should be clear from much of what we have written, deep RL is a work in progress.  In this sense there is also the opportunity for neuroscience research to influence deep RL, continuing the synergistic ‘virtuous circle’ that has connected neuroscience and AI for decades \citep{hassabis2017neuroscience}.

\clearpage
\bibliographystyle{neuron}
\bibliography{references}

\end{document}